\documentclass{article}
\usepackage{ijcai21}

\usepackage{times}

\usepackage{soul}
\usepackage{url}
\usepackage[hidelinks]{hyperref}
\usepackage[utf8]{inputenc}
\usepackage[small]{caption}
\usepackage{subcaption}
\usepackage{graphicx}
\usepackage{amsmath,amssymb}
\usepackage{booktabs}
\usepackage{placeins}
\usepackage{natbib}
\title{Interpretable Mammographic Image Classification using Case-Based Reasoning and Deep Learning}

\def\anonymous{}
\let\anonymous\undefined 

\ifdefined\anonymous
  \author{Keywords: deep learning, interpretability, CNN, mammography, medical, case-based reasoning, CBR, computer vision}
\else
  \author{
    Alina Jade Barnett$^1$\footnote{Contact Author}\and
    Fides Regina Schwartz$^2$\and
    Chaofan Tao$^1$\and
    Chaofan Chen$^3$\and
    Yinhao Ren$^{2,4}$\and
    Joseph Y. Lo$^2$\And
    Cynthia Rudin$^{1,5}$\\
    \affiliations
    $^1$Department of Computer Science, Duke University, USA\\
    $^2$Department of Radiology, Duke University, USA\\
    $^3$School of Computing and Information Science, University of Maine, USA\\
    $^4$Department of Biomedical Engineering, Duke University, USA\\
    $^5$Department of Electrical and Computer Engineering, Duke University, USA\\
    \emails
    \{alina.barnett, fides.schwartz\}@duke.edu,
    chaofan.tao@gmail.com,
    chaofan.chen@maine.edu,
    \{yinhao.ren, joseph.lo\}@duke.edu,
    cynthia@cs.duke.edu
    }
\fi

\begin{document}

\maketitle

\begin{abstract}
When we deploy machine learning models in high-stakes medical settings, we must ensure these models make accurate predictions that are consistent with known medical science. Inherently interpretable networks address this need by explaining the rationale behind each decision while maintaining equal or higher accuracy compared to black-box models. In this work, we present a novel interpretable neural network algorithm that uses case-based reasoning for mammography. Designed to aid a radiologist in their decisions, our network presents both a prediction of malignancy and an explanation of that prediction using known medical features. In order to yield helpful explanations, the network is designed to mimic the reasoning processes of a radiologist: our network first detects the clinically relevant semantic features of each image by comparing each new image with a learned set of prototypical image parts from the training images, then uses those clinical features to predict malignancy. Compared to other methods, our model detects clinical features (mass margins) with equal or higher accuracy, provides a more detailed explanation of its prediction, and is better able to differentiate the classification-relevant parts of the image.
\end{abstract}

\begin{figure*}
    \centering
    \includegraphics[width=\textwidth]{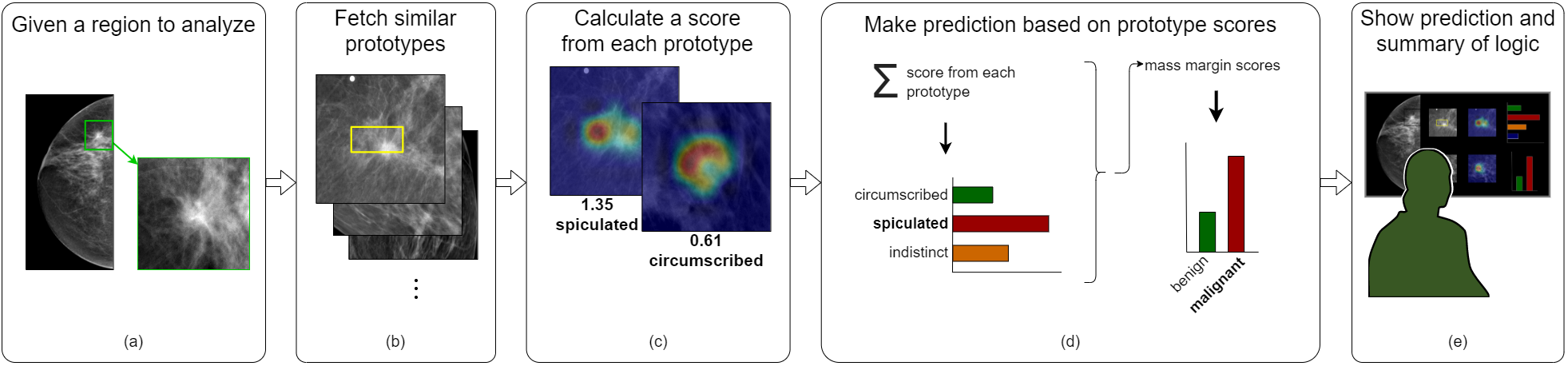}
        \caption{(a,b) Our model compares a new image with its learned set of prototypes. The learned prototypes are each linked to a clinically relevant feature. (c) A similarity score between the new case and each prototype is calculated from the $\ell_2$ distance between the new image patches and the prototypes. (d) The model uses these similarity scores determine the mass margin and the mass malignancy. (e) A radiologist or other domain expert can dismiss the model's prediction if they see incorrect reasoning.}
    \label{fig:overview}
\end{figure*}

\begin{figure*}
    \centering
    \includegraphics[width=0.9\textwidth]{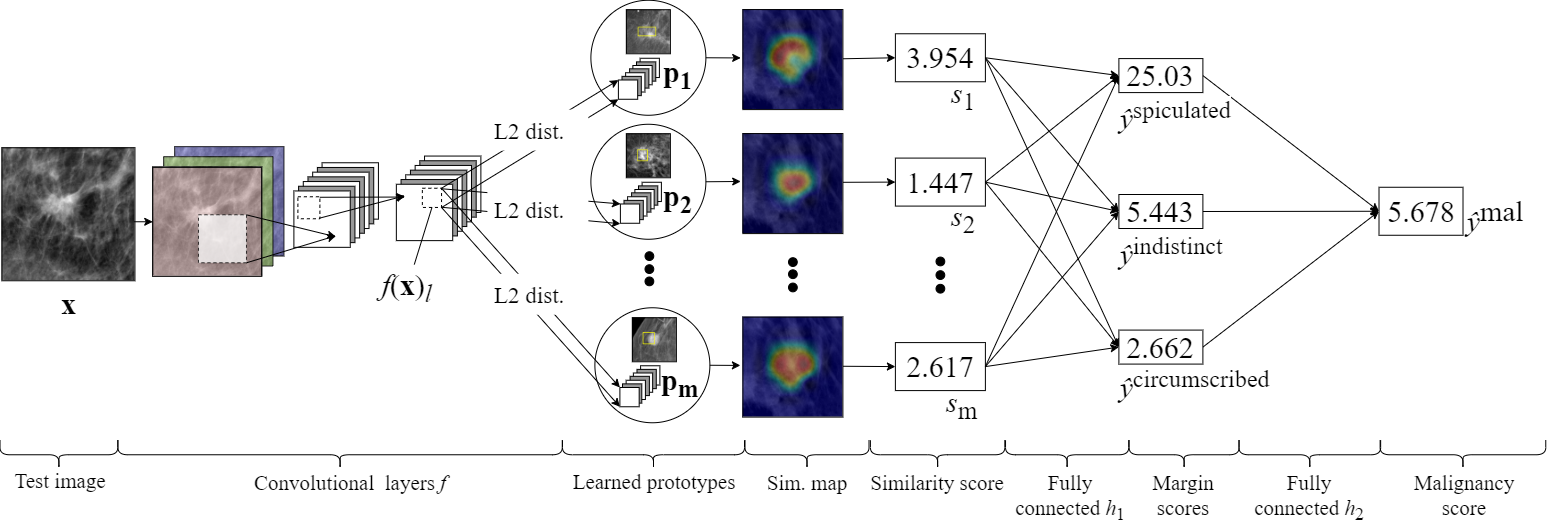}
    \caption{The architecture of our prototype network. Test image $\textbf{x}$ feeds into convolutional layers $f$. Each patch $f(\textbf{x})_l$ of $f(\textbf{x})$ is compared to each learned prototype $\textbf{p}_\textbf{i}$ by calculating the $\ell_2$ distance between the patch and the prototype. The similarity map shows the closest (most ``activated," i.e., smallest $\ell_2$ distance) patches in red and the furthest patches in blue, overlaid on the test image. Similarity score $s_i$ is calculated from the corresponding similarity map. The similarity scores $\textbf{s}$ feed into fully connected layer $h_1$, outputting margin scores $\hat{\textbf{y}}^{\textrm{margin}}$. Margin logits $\hat{\textbf{y}}^{\textrm{margin}}$ feed into fully connected layer $h_2$, outputting malignancy score $\hat{y}^{\textrm{mal}}$.}
    \label{fig:architecture}
\end{figure*}

\begin{figure*}
    \centering
    \includegraphics[width=0.7\textwidth]{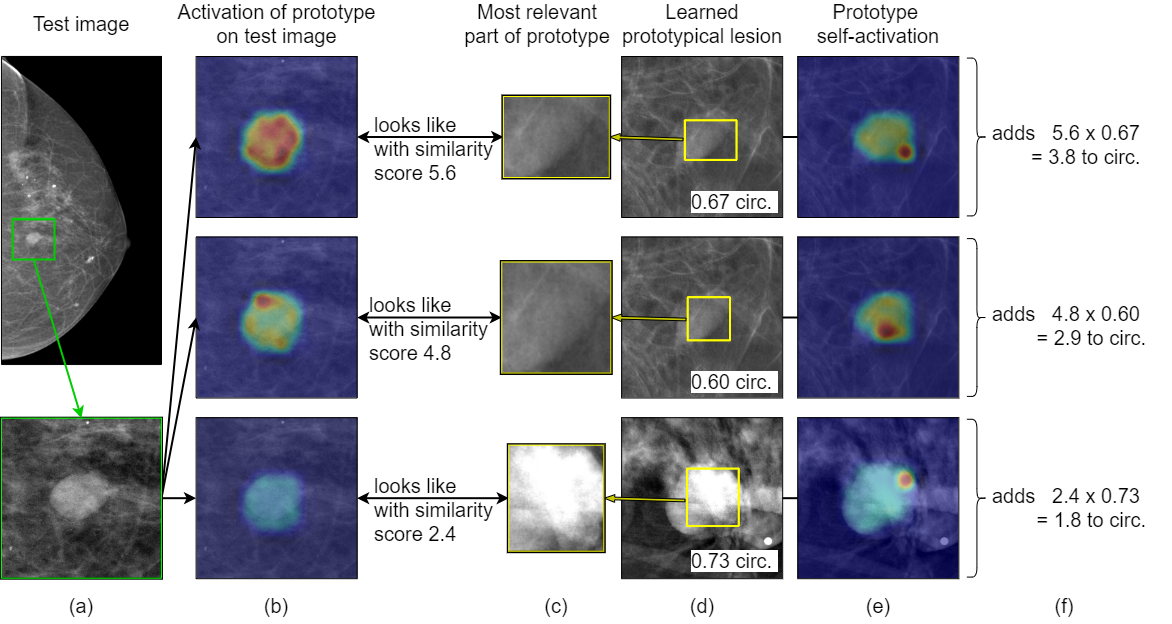}
        \caption{This circumscribed lesion is correctly classified as circumscribed. (a) The region of interest (outlined in green) is fed into the model as the test image. (b) The prototype activation maps for the three closest (most activated) prototypes over the test image. (c) The visual representation of the prototype. (d) The training image which contains the prototypical patch for this prototype. (e) The prototype activation map for this prototype over the image which contains it. (f) The contribution of each prototype to the class score.}
    \label{fig:IAIA_expls}
\end{figure*}

\section{Introduction}

The medical applications of computer vision are broad and promising \citep{choy2018current,hosny2018artificial}. Work that would take a medical specialist hours can be done by a computer algorithm in seconds. Conventional machine learning is currently used for computer-aided detection (i.e., to detect ``lesion'' vs$.$ ``no lesion''), but to make a greater clinical contribution, future approaches need to be able to assist with harder tasks, such as ``should the patient get a biopsy for that lesion?'' These decisions can be challenging, even for the most experienced radiologists, as shown by relatively low inter-rater agreement scores for such difficult tasks \citep{park2007observer,abdullah2009breast,baker1996breast,lazarus2006bi,el2015breast}. 

Interpretability is particularly important for medical applications, where we encounter high stakes and issues of confounding. Neural networks models often use context or confounding information instead of the information that a human would use to solve the same problem in both medical \citep{wang2019removing} and non-medical applications \citep{hu2020stratified}. In some previous studies that used CNNs for medical imaging, researchers created models that appeared to perform well on their test sets, yet upon further inspection, based their decisions on confounding information (e.g., type of equipment) rather than medical information \citep{badgeley2019deep,winkler2019association,zech2018variable}. For high-stakes applications in medicine, model decisions must use relevant medical information rather than context or background information. To ensure clinical acceptance, an AI tool will need to provide its reasoning process to its human radiologist collaborators in order to be a useful aide in these difficult and high-stakes decisions \citep{edwards-fda,soffer2019convolutional}.

Despite their promising performance, deep neural networks are difficult to understand by humans \citep{reyes2020interpretability}. There are two distinct approaches to address this challenge: (1) \textit{Design inherently interpretable networks}, whose reasoning process is constrained to be understandable to humans. (2) \textit{Explain black box neural networks} posthoc by creating approximations, saliency maps, or derivatives. Posthoc explanations can be problematic; for instance \textit{saliency maps} highlight regions of the image (show attention), but can be unreliable and misleading, as they tend to highlight edges and do not show what computation is actually done with the highlighted pixels \citep{rudin2019stop,adebayo2018sanity,arun2020assessing}. There are several types of approaches in interpretable machine learning, including case-based reasoning (which we use here), forcing the network to use logical conditions within its last layers  \citep[e.g.,][]{wu2019towards}, or disentangling the neural network's latent space \citep[e.g.,][]{chen2020concept}. Case-based reasoning models are also used in medicine to retrieve existing similar cases in order to determine how to handle a new case \citep{demigha2004case,macura1995macrad,floyd2000case,kobashi2006computer}.

Unlike existing black-box systems that aim to replace a doctor \citep{mckinney2020international}, we aim to create an interpretable AI algorithm for breast lesions (abbreviated IAIA-BL) whose explicit case-based reasoning can be understood and verified by a medical practitioner. Our novel deep learning architecture enables IAIA-BL to provide an explanation that shows the underlying decision-making process for each case. Figure \ref{fig:overview} shows a schematic of this system: the algorithm compares a new image to a set of previous cases, specifies parts of the new image it considers similar to parts it has seen before, uses that similarity to provide a score for relevant medical features, and uses those scores to predict the likelihood of malignancy. Such a model may be integrated into a clinical support system for classifying breast lesions, because it can point to mammogram regions that resemble prototypical signs of cancerous growth (e.g., spiculated mass margin), and thereby assist doctors in making diagnoses. The radiologist may choose whether to trust IAIA-BL's prediction based not only on the model's measured accuracy, but also based on the plausibility of the explanation generated for the specific case under consideration. Models trained with this architecture provide local interpretability through explanation of each prediction as in Figure \ref{fig:IAIA_expls}, and global interpretability through examination of the set of learned prototypes. To extract more information from our limited dataset, we collected a small set of pixel-level (``fine'') annotations from our radiology team which permitted better generalization using a smaller number of images. We adjusted our algorithm to incorporate a combination of data with whole image labelling and data with pixel-wise annotations, leading to better accuracy and interpretability. 
This novel approach can reduce the confounding in deep learning by leveraging both relatively-abundant coarsely-annotated data and a small amount of finely annotated data. This approach is also practical in the sense that for real-world problems, annotated data are relatively less abundant and more expensive to obtain.

The main contributions of our paper are as follows:
\begin{itemize}
    \itemsep0em 
    \item We developed an inherently interpretable deep-learning model for medical imaging that goes beyond simple attention in its explanations. Our system, IAIA-BL, makes predictions for mammographic breast masses using a case-based reasoning structure by comparing the parts of new images with a learned set of prototypical parts from previous images.
    
    \item We developed a framework for machine learning-based mammography interpretation in line with the goals of radiologists: in addition to predicting whether a lesion is malignant or benign, our work aims to follow the reasoning processes of radiologists in detecting specific clinically relevant aspects of each image, such as the characteristics of the mass margins.
    
    \item We developed a novel training scheme for IAIA-BL which allows it to incorporate prior knowledge in the form of fine-grained expert image annotations. Using only a small number of finely annotated training data and imposing a novel fine-annotation loss on those data, IAIA-BL learns medically relevant prototypes, effectively addresses aspects of confounding issues in medical machine learning models, and sets IAIA-BL apart from ProtoPNet  \citep{PPNet} and other prior works.
\end{itemize}

\section{Methods}
\subsection{Data}
\ifdefined\anonymous
    Our dataset consists of 1136 digital screening mammogram images from 484 patients at [redacted for review] University Health System from 2008 to 2018. These conventional mammography images were collected in compliance with HIPAA under [redacted for review] Health IRB Pro00012010 and waiver of informed consent. The BI-RADS features of mass shape and mass margin were labelled by one fellowship-trained breast imaging radiologist. The ground truth for malignancy of each mass is the result of definitive histopathology diagnosis.
\else
    Our dataset consists of 1136 digital screening mammogram images from 484 patients at Duke University Health System from 2008 to 2018. These conventional mammography images were collected in compliance with HIPAA under Duke Health IRB Pro00012010 and waiver of informed consent. The BI-RADS features of mass shape and mass margin were labelled by a fellowship-trained breast imaging radiologist. The ground truth for malignancy of each mass is the result of definitive histopathology diagnosis.
\fi

The 1136 masses consisted of the following mass margins: 125 spiculated, 220 indistinct, 41 microlobulated, 579 obscured, and 171 circumscribed. We excluded lesions with microlobulated margins because of the small number of lesions represented. We excluded lesions with obscured margins because this margin class is not a good indicator for classifying malignancy, but instead usually indicates the need for follow-up imaging. We split each remaining margin class into 73\% training, 12\% validation, and 15\% testing, ensuring that within each class there was no patient overlap between the testing set and other sets. We performed data augmentation such that each training image is randomly flipped, rotated, and undergoes random cropping with a crop size of 80\% of the image's original size. Each class is augmented to have 5000 images for the training set.

\subsection{Model Architecture}
Figure \ref{fig:architecture} gives an overview of our model architecture. 

Given a region of interest $\mathbf{x}$ in a mammogram, our IAIA-BL model first extracts useful features $f(\mathbf{x})$ for mass-margin classification, using a series of convolutional layers $f$ from a VGG-16 network \citep{simonyan2015very} pre-trained on ImageNet. The output $f(\mathbf{x})$ is size $14 \times 14 \times c$, where $c$ is the number of channels. $l \in \{(1,1),...,(1,14),(2,1),...,(14,14)\}$ indexes each of the $1 \times 1 \times c$ patches $f(\mathbf{x})_l$ across the spatial dimensions.

Following the convolutional layers $f$ is prototype layer $g$. As in ProtoPNet \citep{PPNet}, the prototype layer contains $m$ prototypes $\mathbf{P}=\{\mathbf{p}_j\}_{j=1}^m$ learned from the training set. Each prototype is size $1 \times 1 \times c$. Since a prototype has the same number of channels but a smaller spatial dimension than the convolutional feature maps $f(\mathbf{x})$, we can interpret the prototype as representing a prototypical activation pattern of its class and we can visualize the prototype as a patch of the training image it appears in. 

Given convolutional feature maps $f(\mathbf{x})$, the prototype layer $g$ calculates the similarity score $s_{j}$ between image $\mathbf{x}$ and each prototype $\mathbf{p}_j$. It first computes the distance between $\mathbf{p}_j$ and each of the $l$ $1 \times 1$ spatial patches of convolutional feature map $f(\mathbf{x})$ by:
$
    d_{j,l} = \|\mathbf{p}_j-f(\mathbf{x})_l\|_2^2,
$
and converts distances to similarities:
\begin{align}
    s_{j,l} =& \log\frac{d_{j,l}+1}{d_{j,l}+\epsilon}.
\end{align}
This provides a set of similarity scores $\{ s_{j,l} \}_{l=(1,1)}^{(14,14)}$ which can be arranged spatially into a similarity map $[s_{j,l}]_{l=(1,1)}^{(14,14)}$ comparing the input image and each prototype $\mathbf{p}_j$. The overall similarity score $s_j$ is calculated using top-k average pooling \citep[as in][]{kalchbrenner2014convolutional}:
\begin{align}
    s_{j} =&~ \textrm{avg} \left( \textrm{top}_k \left( \{ s_{j,l} \}_{l=(1,1)}^{(14,14)} \right) \right).
\end{align}

Conceptually, this means that if the input image $\mathbf{x}$ has spicules along the mass margin, its convolutional feature maps $f(\mathbf{x})$ will have patches $f(\mathbf{x})_l$ that represent the spicules from the input image. These patches will be close (in $\ell_2$ distance in the latent space) to one or more prototypes $\mathbf{p}_j$ that represent spicules on the mass margin -- consequently, the similarity scores $s_{j,l}$ between those spiculated prototypes and those spiculated patches will be large.

In IAIA-BL, we initialize the model with $m=15$. We prune duplicate prototypes, and the final IAIA-BL model presented has 4 prototypes for circumscribed mass margin, 3 prototypes for indistinct mass margin, and 4 prototypes for spiculated mass margin. We set $c$ to $512$ in our experiments.

IAIA-BL ends with two fully connected layers. The first fully connected layer $h_1$ multiplies the vector of similarity scores $[s_1, ..., s_m]$ by a weight matrix to produce three output scores $\hat{y}^{\text{circumscribed}}$, $\hat{y}^{\text{indistinct}}$, and $\hat{y}^{\text{spiculated}}$, one for each margin type. These are (afterwards) normalized using a softmax function to generate the probabilities that the mass margin in the input image belongs to each of the three mass-margin types. The second fully connected layer $h_2$ then combines the vector of (unnormalized) mass-margin scores $\hat{\mathbf{y}}^{\text{margin}}=[\hat{y}^{\text{circumscribed}}, \hat{y}^{\text{indistinct}}, \hat{y}^{\text{spiculated}}]$ into a final score of malignancy $\hat{y}^{\text{mal}}$, which is passed into a logistic sigmoid function to produce a probability that the input image has a malignant breast cancer.

\subsection{Model Training} \label{sec:model_training}

The training of IAIA-BL differs from that of ProtoPNet \citep{PPNet} in three ways: (1) IAIA-BL was trained with a fine-annotation loss which penalizes prototype activations on medically irrelevant regions for the subset of data with fine annotations. (2) IAIA-BL considers the top $5\%$ of the most activated convolutional patches that are closest to each prototype, instead of only the top most activated patch. (3) We include an additional fully connected layer to transform mass margin scores $\hat{\mathbf{y}}^{\text{margin}}$ to malignancy score $y^{\text{mal}}$ whose training is isolated from the rest of the network. 

We represent the dataset of $n$ training images $\mathbf{x}_i$, with mass-margin labels $y^{\text{margin}}_i$ and malignancy labels $y_i^{\text{mal}}$, as $D=\{(\mathbf{x}_i, y^{\text{margin}}_i, y_i^{\text{mal}})\}_{i=1}^n$. 30-image subset $D' \subseteq D$ comes with the radiologist's (fine) annotations of where medically relevant information is in that training image. For a training instance $(\mathbf{x}_i, y^{\text{margin}}_i, y_i^{\text{mal}}) \in D'$, we define a fine-annotation mask $\textbf{m}_i$, such that $\textbf{m}_i$ takes the value $0$ at those pixels that are marked as ``relevant to mass margin identification,'' and takes the value $1$ at other pixels. Each fine-annotation mask $\textbf{m}_i$ has the same spatial dimensions (height and width) as the training image $\textbf{x}_i$.

The training of IAIA-BL has four stages: (A1) training of the convolutional layers $f$ and the prototype layer $g$; (A2) projection of prototypes; (A3) training of the first fully connected layer $h_1$ for predicting mass-margin types; and (B) training of the second fully connected layer $h_2$ for predicting malignancy probability. Stages A1, A2, and A3 are repeated until the training loss for predicting mass-margin types converges, then we move to Stage B. 

    
\textbf{Stage A1:} In the first training stage, we aim to learn meaningful convolutional features. In particular, we want convolutional features that represent a particular mass-margin type to be clustered around a prototype of that particular mass-margin type, and to be far away from a prototype of other mass-margin types. As in \citet{PPNet}, we jointly optimize the parameters $\mathbf{\theta}_f$ of $f$, and the prototypes $\mathbf{p}_1$, ..., $\mathbf{p}_m$ of $g$, while keeping the two fully connected layers $h_1$ and $h_2$ fixed. We minimize the following training loss:
\begin{align}
    \textrm{min}_{\mathbf{\theta}_f, \mathbf{p}_1, ..., \mathbf{p}_m} &\textrm{CrsEnt} + \lambda_c \textrm{Clst} + \lambda_s \textrm{Sep} + \lambda_f \textrm{Fine}, 
    \label{eq:joint_training_obj}
\end{align}
where the cross entropy term 
penalizes  misclassification of mass-margin types on the training data
. The cross entropy term also ensures that the learned convolutional features and the learned prototypes are relevant for predicting mass-margin types. 

Differing from \citet{PPNet} by the use of $\mathrm{mink}$ instead of $\min$, the cluster and separation cost are defined by:
\begin{align}
    \textrm{Clst} =& \frac{1}{n}\sum_{i=1}^{n} \min_{j:\text{class}(\mathbf{p}_j)=y^{\text{margin}}_i} \left(\gamma \right), \text{  and  } \label{eq:cluster_cost}\\
    \textrm{Sep} =& -\frac{1}{n}\sum_{i=1}^{n} \min_{j:\text{class}(\mathbf{p}_j) \neq y^{\text{margin}}_i} \left(\gamma \right), \text{  with  } \label{eq:separation_cost}\\
    \gamma =& \frac{1}{k}\sum\mathrm{mink}_{\mathbf{z} \in \text{patches}(f(\mathbf{x}_i))}\left(\|\mathbf{z}-\mathbf{p}_j\|_2^2\right)
\end{align}
where $\mathrm{mink}$ gives the $k$ smallest squared distances. 
Empirically, we found that IAIA-BL trained with the relaxed cluster and separation costs outperforms the one trained with the original (i.e., $k=1$) cluster and separation costs of \citet{PPNet} on the task of margin classification, possibly because the relaxed cluster and separation cost (along with the top-$k$ average pooling) allows the gradient of the loss function to back-propagate through $k$ convolutional patches, instead of just $1$ patch, during training -- consequently, the gradient will be \textit{less sensitive} and \textit{more robust} to changes in the location of the most activated convolutional patch by each prototype.

\begin{figure}
    \centering
    \includegraphics[width=0.7\linewidth]{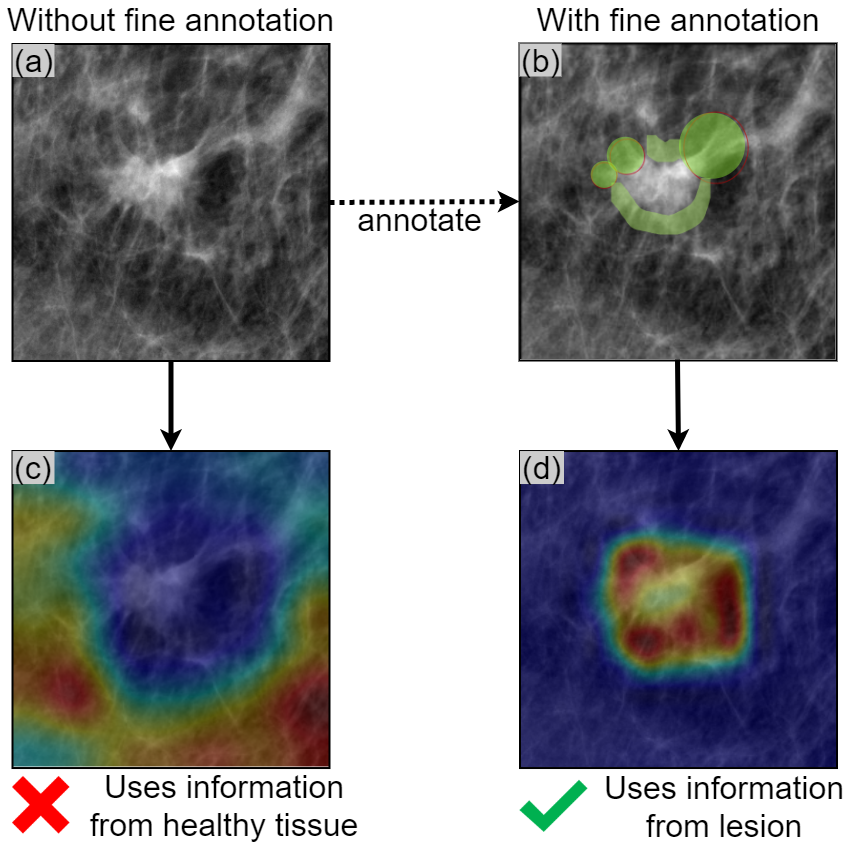}
    \caption{By introducing a constraint on model attention, we penalize the model for using confounding information. (a) Shows the lesion to be classified. (b) The spicules of the lesion have been marked in green by a radiologist. (c) Without fine annotation loss in training, the activation map highlights confounding information. (d) Using fine annotation loss during training, the activation map highlights relevant information -- areas that contain spicules. The attention is within the area marked by the radiologist (otherwise it would have been penalized by the fine-annotation loss function).} \label{fig:fine_attention}
\end{figure}

The fine-annotation loss, which is new to this paper, penalizes prototype activations on medically irrelevant regions of radiologist-annotated training mammograms; see Figure \ref{fig:fine_attention}. The fine-annotation loss is defined by:
\begin{align}
\textrm{Fine} = \sum_{i \in D'}  \Bigg(\sum_{j: \text{class}(\mathbf{p}_j) = y^{\text{margin}}_i} \|\mathbf{m}_i \odot \textrm{PAM}_{i,j}\|_2
\nonumber \\
+ \sum_{j: \text{class}(\mathbf{p}_j) \neq y^{\text{margin}}_i} \|\textrm{PAM}_{i,j}\|_2\Bigg)
\end{align}
where \textit{prototype activation map} $\textrm{PAM}_{i,j}$ for prototype $\mathbf{p}_j$ over image $\textbf{x}_i$ is computed by bilinearly upsampling the similarity map $[s_{j,l}]_{l=(1,1)}^{(14,14)}$ to yield the the same dimensions (height and width) as the fine-annotation mask. This promotes the learning of prototypes that stay away from any features that could appear in classes that are not the prototypes' designated classes, so that the prototypes of a particular class represent distinguishing features of that class.

To incorporate the training data with fine annotations into model training, we construct batches with $75$ training examples from $D$ with lesion-scale annotations and $10$ training examples from $D'$ with fine-scale annotations. The fine-annotation loss on a lesion-scale annotation penalizes activation outside of the area marked as the lesion, whereas the fine-annotation loss on a finely annotated image penalizes activation outside of the region ``relevant to the mass margin class'' as marked by the radiologist. 

The prototype layer was initialized randomly using the uniform distribution over a unit hypercube (because the convolutional features from the last convolutional layer all lie between $0$ and $1$). In our experiments, $\lambda_c = 0.8$, $\lambda_s = 0.08$, and $\lambda_f = 0.001$. 

\textbf{Stage A2:} As in work of \citet{PPNet}, we project the prototypes $\textbf{p}_j$ onto the nearest convolutional feature patch from the training set $D$, of the same class as $\textbf{p}_j$. 

\textbf{Stage A3:} 
In this stage, we fine-tune the first fully connected layer $h_1$ to further increase the accuracy in predicting mass-margin types. We fix the parameters $\mathbf{\theta}_f$ and the prototypes $\mathbf{p}_1$, ..., $\mathbf{p}_m$, and minimize the following training objective with respect to the parameters $\mathbf{\theta}_{h_1}$ of the first fully connected layer $h_1$:
\begin{align}
\textrm{min}_{\mathbf{\theta}_{h_1}} \frac{1}{n} \sum_{i=1}^n \textrm{CrossEntropy}(h_1 \circ g \circ f(\mathbf{x}_i), y^{\text{margin}}_i). \label{eq:h1_training}
\end{align}

The first time we enter stage A3, we initialize connections in fully connected layer $h_1$ to a value of 1 for prototypes that are positive for that mass margin, -1 otherwise.


\textbf{Stage B:} In this stage, we train the second fully connected layer $h_2$ for predicting malignancy probability, using a logistic regression model whose input is the (unnormalized) mass-margin scores produced by the first fully connect layer $h_1$, and whose output is the probability of malignancy. To prevent the malignancy information from biasing the mass margin classification, we train the model in a modular style and it is not trained completely end-to-end in any stage, i.e., there is no return to Stage A from Stage B.

\section{Experiments}
\subsection{Mass Margin Classification} \label{sec:mass_margin_exps}
We compare the following models.

    \textbf{IAIA-BL.} The final model is trained on the union of the training set and validation set, and tested on a test set never before seen in training. We trained for 50 epochs because model training generally converges between 40 and 50 epochs. We selected hyperparameters $k=5$ and $\lambda_f=0.001$ by testing over the validation set. Our model can be fully trained on one P100 GPU in 50 hours.

    \textbf{ProtoPNet} \citep{PPNet} has no fine annotation loss, and uses max pooling logic where IAIA-BL uses top-$k$ average pooling logic.

    \textbf{VGG-16 with GradCAM and GradCAM++} \citep{simonyan2015very,Selvaraju_2017_ICCV,chattopadhay2018grad} is trained with two added fully connected layers to account for the larger number of parameters in our model. Pre-trained on ImageNet, it was trained for 250 epochs and the epoch with the highest test accuracy was selected for comparison. There is no native way to incorporate our fine annotation into VGG-16. Because VGG-16 provides no inherent interpretability or localization, we used the posthoc GradCAM and GradCAM++ techniques to generate localization information and calculate activation precision. 
    
\subsubsection{Metrics} \label{subsec:activprec}
We use the average of the AUROCs (area under receiver operator characteristic curve) for each of the three mass margin classes as the performance metric for mass-margin prediction. 95\% confidence intervals were derived using Delong's method \citep{delong1988comparing,sun2014fast}.

We designed the interpretability metric \textit{activation precision} to quantify what proportion of the information used to classify the mass margin comes from the ``relevant region'' as marked by the radiologist-annotator. The activation precision for $\mathbf{x}_i$ is defined as:
\begin{align}
    \sum_{j: \text{class}(\mathbf{p}_j)=y^{\text{margin}}_i}
    \left(\frac{\sum \left[(1 - \mathbf{m}_i) \odot T_{\tau}\left(\textrm{PAM}_{i,j}\right)\right]}{\sum T_{\tau}\left(\textrm{PAM}_{i,j}\right)}\right), \label{eq:ap}
\end{align}
where $T_{\tau}$ is a threshold function that returns the top $(1-\tau) \times 100\%$ of the input values as $1$ and the bottom $\tau \times 100\%$ as $0$. The fraction in (\ref{eq:ap}) gives a proportion of highly activated pixels that are medically relevant. To evaluate activation precision for GradCAM and GradCAM++, we calculate as in (\ref{eq:ap}) but replace the prototype activation map $\textrm{PAM}_{i,j}$ with the normalized gradient map for the correct class.

Activation precision can be measured both at \textit{lesion-scale} (i.e., is the activation within the lesion area and not the surrounding healthy tissue?) and at \textit{fine-scale} (i.e., is the activation on the specific part of the margin marked relevant by the radiologist?). 

Activation precision is a measure of interpretability, in the sense that the higher the activation precision, the better a prototype is at detecting medically relevant features for mass-margin classification. $\tau = 0.95$. 
95\% confidence intervals were derived using non-parametric bootstrap resampling with 5000 samples each equal to the size of the test set.

\subsubsection{Results for Mass Margin Classification} \label{sec:results_margin}

\begin{table}
\centering
\begin{tabular}{lll}
\hline
Model  & Avg. AUROC & Act. Prec. \\
\hline
IAIA-BL (ours)  & \textbf{0.951} [0.9, 1.0]  & \textbf{0.94} [0.9, 1.0]    \\
ProtoPNet & 0.911 [0.8, 1.0]  & 0.51 [0.3, 0.7]    \\
VGG-16 &  0.947 [0.9, 1.0]  &   ---   \\
~~with GradCAM & 0.947 [0.9, 1.0]  & 0.45 [0.4, 0.5]     \\
~~with GradCAM++ &  0.947 [0.9, 1.0] & 0.53 [0.4, 0.6]    \\
\hline
\end{tabular}
\caption{For the task of mass margin classification, our model achieves equal or better AUROC (performance) and significantly better lesion-scale activation precision (interpretability). The 95\% confidence interval is shown after each value.}
\label{tab:results_sum}
\end{table}

The full results table can be found in Appendix \ref{app:roc_curves} with summary results shown in Table \ref{tab:results_sum}.

IAIA-BL's AUROC for mass margin classification is as good or better than the AUROCs of interpretable ProtoPNet and the analogous black-box model, VGG-16. Both ProtoPNet and VGG-16 show significantly lower activation precision than IAIA-BL. \textit{Both baselines use information from image regions entirely outside the region that contains the lesion.} The baseline models are not restricted from using confounding information, and thus do so freely. \textit{These models should not be used in practice for this reason.} A visual comparison of activation maps is shown in Appendix \ref{app:viz_compare}.

The Cohen $\kappa$ metric measures agreement between two sets of labels, where a higher value represents better agreement. With a $\kappa$ value of 0.74 (95\% CI: 0.60, 0.86), IAIA-BL's predictions on the test set and our physician-annotator's labels have ``substantial'' agreement \citep{landis1977application}. The agreement between our model and our physician-annotator is greater than the agreement of physicians with each other, e.g., 0.61-0.65 in \citet{baker1996breast}, 0.58 in \citet{RAWASHDEH2018294}, and 0.48 in \citet{lazarus2006bi}.

\subsection{Malignancy Prediction}

\begin{table}
\centering
\begin{tabular}{ll}
\hline
Model  & AUROC \\
\hline
IAIA-BL (ours)  & 0.84 [0.74, 0.94]\\
Radiologists' Estimates & 0.91 [0.85, 0.97]\\
End-to-end VGG-16 & 0.87 [0.82, 0.93]\\
\hline
\end{tabular}
\caption{For the task of malignancy prediction. The 95\% confidence interval is shown after each value. Here, VGG-16 is permitted to use confounding information. IAIA-BL is restricted only to use mass margin information to make its prediction.}
\label{tab:mal_results}
\end{table}

IAIA-BL converts mass margin scores $\hat{\textbf{y}}^{\textrm{margin}}$ to malignancy score $\hat{y}_i^{\text{mal}}$ with the following concise linear model learned in training stage B:
\begin{align}
    \hat{y}_i^{\text{mal}} = -16~ \hat{y}_i^{\text{circumscribed}} -10 ~\hat{y}_i^{\text{indistinct}} + 6~ \hat{y}_i^{\text{spiculated}}.
\end{align}
$\text{Prob}(\text{malignancy}) = \sigma((\hat{y}_i^{\text{mal}} - 155) / 100 )$, where $\sigma(t)$ is the logistic sigmoid function.

This interpretable model is consistent with known medical knowledge: a high spiculated score results in a high probability of malignancy, while high circumscribed or indistinct margin scores indicate a benign lesion.

\textbf{Radiologists' estimates.} During data collection, we asked radiologists to estimate the probability that the lesion will be malignant. There are several caveats for this estimate: radiologists do not perform this task in standard practice; and the annotations were completed as part of a separate study that used consumer-grade monitors without the necessary specifications or calibrations of medical-grade displays. Nonetheless, these estimates represent the radiologist's ``best guess'' for these data.

\textbf{End-to-end VGG-16.} This baseline is uninterpretable VGG-16 trained to predict malignancy, but not restricted to predicting from only the mass margin results. As with the mass margins, VGG-16 likely uses confounding information; e.g., the age of the patient could be inferred from the density of the normal breast tissue and could be a useful predictor of malignancy. 

\subsubsection{Results for Malignancy Prediction} \label{sec:results_mal}

As shown in Table \ref{tab:mal_results}, \textit{even though IAIA-BL is constrained to predict malignancy using only the mass margin scores} $\hat{\textbf{y}}^{\textrm{margin}}$ (rather than extra information that may be contained within the raw pixels of the image), it performs comparably to uninterpretable end-to-end VGG-16. If our dataset were larger, and if we include non-imaging features such as patient age, it could potentially boost performance. 

Overall, the result shows the potential of fully-interpretable deep learning models for mammography, which, prior to this paper, had not been established.

\section{Conclusions}
We were able to create an interpretable mass margin prediction model with equal or higher performance to its uninterpretable counterparts, despite the fact that the uninterpretable models often used confounding information to boost performance. Our model's interpretability, measured by how well its attention agreed with a radiologist annotator's hand-drawn attention maps, exceeded that of existing methods and does not resort to post-hoc analysis.
Using only a small dataset, we were able to provide an interpretable network that performs comparably with radiologists on mass margin classification and malignancy prediction.

\bibliographystyle{named}
\bibliography{ijcai21}

\onecolumn
\appendix
\FloatBarrier
\section{Visual Comparison of Activation Maps} \label{app:viz_compare}
\begin{figure}[h]
    \centering
    \includegraphics[width=\textwidth]{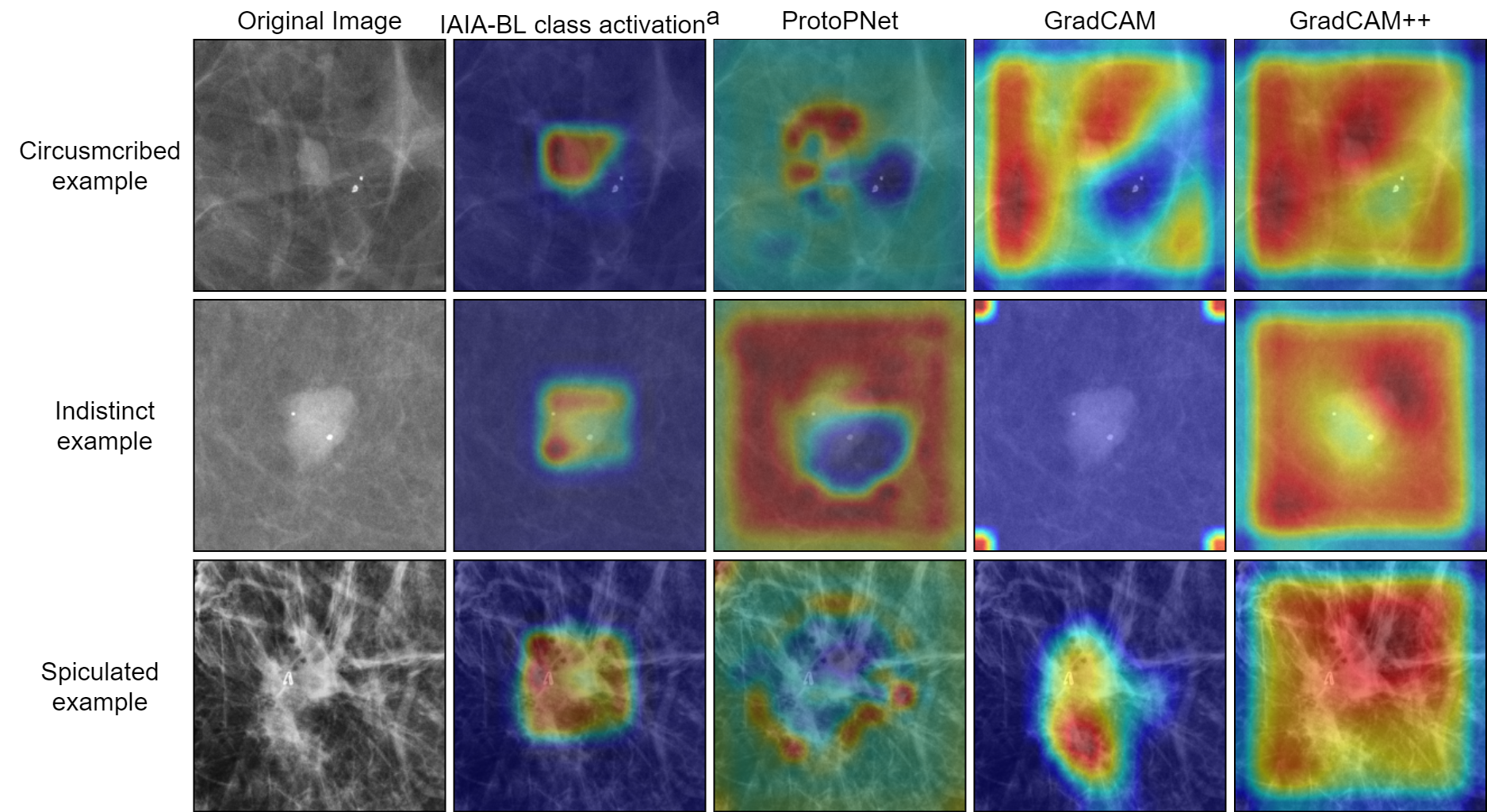}
        \caption{We compare explanations from two common saliency methods (GradCAM and GradCAM++ \protect\citep{Selvaraju_2017_ICCV,chattopadhay2018grad}) to a class activation visualization derived from our method. The explanations from IAIA-BL are more likely to highlight the lesion and less likely to highlight the surrounding healthy tissue. This is shown quantitatively by the activation precision metric defined in Section \ref{subsec:activprec}. $^{\text{a}}$This single image visualization is a dramatic simplification of the full explanation that is generated by IAIA-BL. The IAIA-BL and ProtoPNet class activation visualizations shown in this figure are generated by taking the similarity score-weighted average of prototype activation maps for all prototypes of the correct class. } \label{fig:compare_gradcam_viz_to_us}
\end{figure}

\FloatBarrier
The IAIA-BL and ProtoPNet class activation visualizations shown in Figure \ref{fig:compare_gradcam_viz_to_us} are produced by taking the weighted average of the prototype activation map $\textrm{PAM}_{i,j}$ for every prototype of the correct class. The weight for each prototype is the similarity score between the prototype and the original image. $\textrm{CAV}^{a,b}$, the value of the $a$-th row and $b$-th column of class activation visualization $\textrm{CAV}$, is defined as:

\begin{align}
    \textrm{CAV}^{a,b}(\mathbf{x}_i, y_i^{\text{margin}}) \backsim & \sum_{j: \text{class}(\mathbf{p}_j) = y^{\text{margin}}_i} s_j \left(\textrm{PAM}_{i,j}\right)^{a,b}
\end{align}
where $\textrm{PAM}_{i,j}$ is the prototype activation map for prototype $p_j$ over image $\mathbf{x}_i$. $\textrm{CAV}$ will have the same dimension as $\textrm{PAM}_{i,j}$. $\textrm{CAV}$ is normalized using mix-max normalization so that its values fall between 0 and 1.

Compared to the class activation visualizations produced by baselines with similar predictive performance, the prototype activation maps produced by IAIA-BL are more likely to highlight the lesion and more likely to highlight the relevant part of the mass margin. This is shown quantitatively by the activation precision metric results from Section \ref{sec:mass_margin_exps}.

\FloatBarrier
\section{Results} \label{app:roc_curves}
\FloatBarrier
Table \ref{tab:full_margin_results} shows margin classification test results for models from Section \ref{sec:mass_margin_exps}. The table shows that IAIA-BL's test AUROC performance with respect to all tasks is approximately as good as the best of the baselines. IAIA-BL's main advantage (interpretability) is shown in the bottom two rows of the table, where there is a huge drop in fine activation precision for original ProtoPNet and VGG-16 as compared with IAIA-BL. VGG-16 has no inherent interpretability but posthoc GradCAM and GradCAM++ provide localization information on which we measure activation precision. 
\setlength\tabcolsep{5 pt}
\begin{table}[h]
\small
  \caption{Margin classification results for models from Section \ref{sec:mass_margin_exps}. The first five rows measure prediction performance, whereas the lower two rows measure interpretability performance. For each row, the best value is in \textbf{bold}, and values not significantly different than the best are in \textit{italics}. $^{\textrm{a}}$ Because this technique is posthoc, there is no guarantee that the generated explanation matches the model's decision making.}
  \label{tab:full_margin_results}
  \centering
  \begin{tabular}{llllll}
    \hline
     & \multicolumn{4}{c}{Model} \\
     &  &  & VGG-16 & VGG-16  \\
     & IAIA-BL  & ProtoPNet & with GradCAM & with GradCAM++ \\
    \hline
    Performance (AUROC) &  &  &  & \\
    ~~Mass Margin Class. & \textbf{0.951} [0.905, 0.996]  & \textit{0.911} [0.848, 0.974] & \textit{0.947} [0.898, 0.996] & \textit{0.947} [0.898, 0.996] \\
    ~~~~Spiculated vs. all & \textit{0.96} [0.90, 1.00] & \textbf{0.97} [0.93, 1.00] & \textit{0.95} [0.89, 1.00] & \textit{0.95} [0.89, 1.00] \\
    ~~~~Indistinct vs. all & \textit{0.93} [0.88, 0.99] & \textit{0.87} [0.78, 0.94] & \textbf{0.94} [0.89, 0.99] & \textbf{0.94} [0.89, 0.99] \\
    ~~~~Circumscribed vs. all & \textbf{0.97} [0.94, 1.00] & \textit{0.93} [0.87, 1.00] & \textit{0.95} [0.91, 1.00] & \textit{0.95} [0.91, 1.00]\\
    Cohen's $\kappa$ & \textbf{0.74} [0.60, 0.86] & \textit{0.64} [0.49, 0.78] & \textbf{0.74} [0.60, 0.87] & \textbf{0.74} [0.60, 0.87]\\
\hline
    Interpretability &  &    & &  \\
    ~~Fine-scale Act. Prec. & \textbf{0.41} [0.39, 0.45] & 0.24 [0.17, 0.31] & 0.21 [0.05, 0.43]$^{\textrm{a}}$ & 0.24 [0.08, 0.45]$^{\textrm{a}}$ \\
    ~~Lesion-scale Act. Prec. &  \textbf{0.94} [0.92, 0.97] & 0.51 [0.34, 0.68] & 0.45 [0.37, 0.54]$^{\textrm{a}}$ & 0.53 [0.44, 0.61]$^{\textrm{a}}$ \\
    \hline
  \end{tabular}
\end{table}

            
            
\FloatBarrier
\section{Activation Precision} \label{app:ap_def}
\begin{figure}[h]
        \begin{center}
            \includegraphics[width=\linewidth]{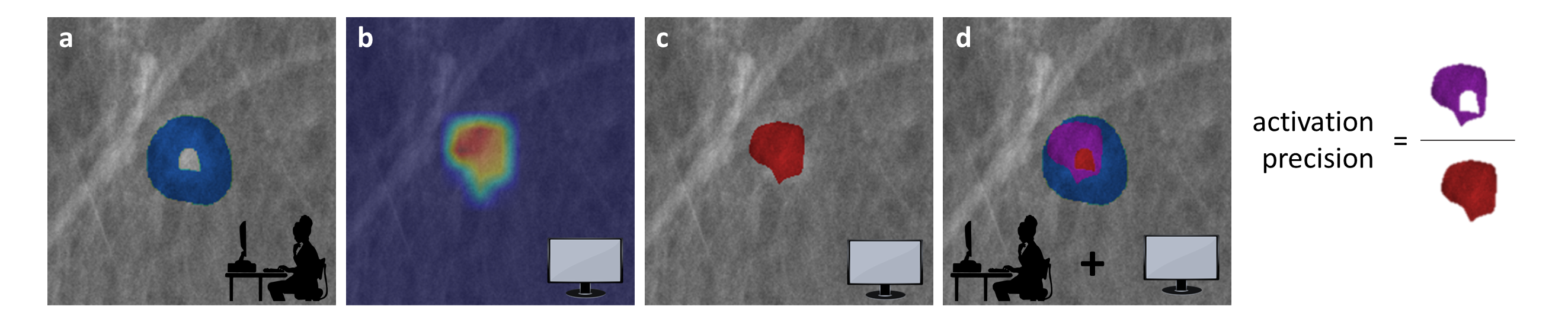}
        \end{center}
        \vfill
          \caption{The definition for activation precision definition given as a visual example. (a) The physician-annotated area of the image, indicating the medically relevant region for mass-margin classification, is highlighted in blue. (b) The activation of a same-class prototype (learned in model training) on the lesion. The red region is most highly activated. (c) A mask of the top activation from \textit{b}. (d) The area of fine annotation from the radiologist annotator is shown in blue as in \textit{a}, the area most activated by a same-class prototype is shown in red as in \textit{c}, the overlap in these two regions is shown in purple.}
        \label{fig:ap_def}
\end{figure}

\FloatBarrier
A visual explanation of the activation precision metric is given in Figure \ref{fig:ap_def}.

Note that we do not compute the proportion of medically relevant pixels that are highly activated, i.e., the denominator in (\ref{eq:ap}) is \textit{not} the number of medically relevant pixels (given by $\sum (1 - \mathbf{m}_i)$). This is because we do not require each prototype to detect the entire mass margin (that was annotated by a doctor), but rather, we expect each prototype to detect a differentiating feature that may only be present at \textit{parts} of a mass margin. Since a prototype may correctly activate on only parts of the margin, intersection over union or measuring the proportion of medically relevant pixels on which the prototype activates highly would not be appropriate metrics.

\end{document}